\pdfoutput=1
\documentclass[11pt]{article}

\usepackage{acl}

\usepackage{times}
\usepackage{latexsym}
\usepackage{amssymb}
\usepackage{amsmath}
\usepackage{graphicx}
\usepackage{booktabs}
\usepackage[T1]{fontenc}
\usepackage{enumitem}

\usepackage[utf8]{inputenc}

\usepackage{microtype}

\usepackage{inconsolata}

\title{Key ingredients for effective zero-shot cross-lingual \\knowledge transfer in generative tasks}

\author{Nadezhda Chirkova \\
  Naver Labs Europe  \\
   Grenoble, France  \\
  \texttt{nadia.chirkova} \\
  \texttt{@naverlabs.com} \\\And
  Vassilina Nikoulina \\
  Naver Labs Europe  \\
   Grenoble, France  \\
  \texttt{vassilina.nikoulina} \\
  \texttt{@naverlabs.com} \\}

\begin{document}
\maketitle
\begin{abstract}
Zero-shot cross-lingual knowledge transfer enables a multilingual pretrained language model, finetuned on a task in one language, make predictions for this task in other languages. While being broadly studied for natural language understanding tasks, the described setting is understudied for generation. Previous works notice a frequent problem of generation in a wrong language and propose approaches to address it, usually using mT5 as a backbone model.
In this work we compare various approaches proposed from the literature in unified settings, also including alternative backbone models, namely mBART and NLLB-200. 
We first underline the importance of tuning learning rate used for finetuning, which helps to substantially alleviate the problem of generation in the wrong language. Then, we show that with careful learning rate tuning, the simple full finetuning of the model acts as a very strong baseline and alternative approaches bring only marginal improvements. Finally, we find that mBART performs similarly to mT5 of the same size, and NLLB-200 can be competitive in some cases. Our final zero-shot models reach the performance of the approach based on data translation which is usually considered as an upper baseline for zero-shot cross-lingual transfer in generation.
\end{abstract}

\section{Introduction}
Multilingual pretrained language models (mPLMs) such as mBERT~\cite{mbert}, mBART~\cite{mbartpt}, and mT5~\cite{mt5} provide high-quality representations for texts in various languages and serve as a a universal backbone for finetuning on language-specific task data. The latter, however, is not always available for a language of interest, providing motivation for studying \textit{zero-shot cross-lingual} capabilities of mPLMs. In this setting, the model is adapted, e.g. finetuned, on input-output pairs in a \textit{source} language, usually English, and then applied in a zero-shot manner to make predictions for inputs in another \textit{target} language, seen only at the pretraining stage. 

\begin{figure}[t!]
    \centering
         \includegraphics[width=\linewidth]{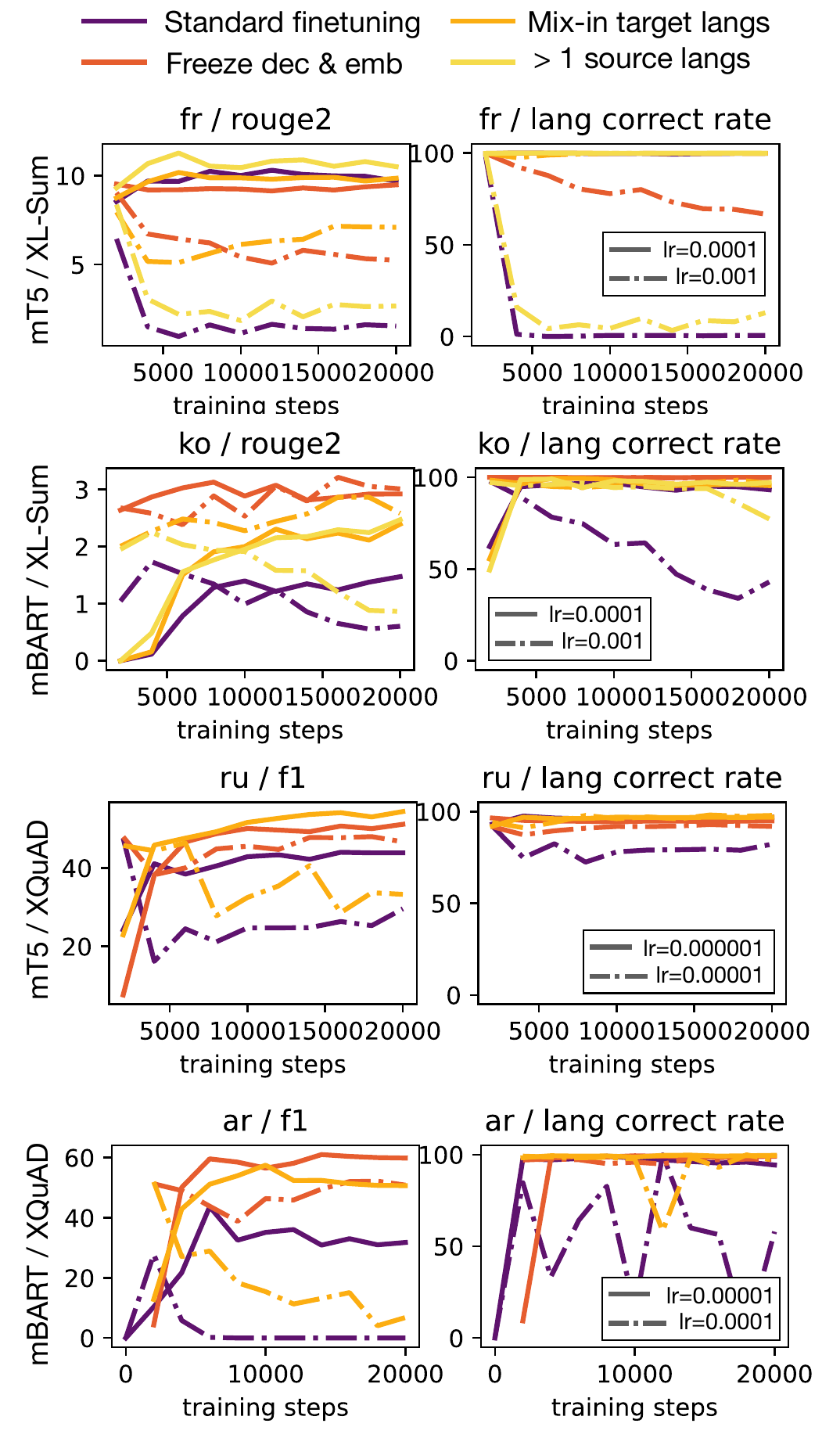}
        \caption{Learning rate plays a key role in cross-lingual transfer: decreasing LR almost completely eliminates generation in the wrong language with standard full finetuning, 
        and often brings larger improvements that using complex adaptation methods developed to overcome this problem.
        Full results in Fig.~\ref{fig:mt5_xlsum}--\ref{fig:mbart_pt_xquad} in Appendix.}
        \label{fig:lr_ft}
\end{figure}

While the described setting was broadly studied for natural language understanding tasks~\cite{mt5,commoncrawl,artetxe-etal-2020-cross,pires-etal-2019-multilingual,wu-dredze-2019-beto,pfeiffer-etal-2020-mad}, work on zero-shot cross-lingual transfer in \textit{generation} is more limited~\cite{vu-etal-2022-overcoming,mmt5,zmbart,li-murray-2023-zero}. Previous work highlight two main problems arising in this scenario:
producing incoherent or irrelevant answers, and generating text in a wrong language. A series of potential solutions were proposed, such as freezing the parts of the weights during finetuning, utilizing parameter-efficient finetuning methods, mixing-in the unsupervised target language data together with the supervised source language data, or using more than one source language. A common strategy is also to perform an intermediate tuning of the model on the language generation task in a self-supervised manner (as opposed to denoising tasks used for pretraining).  

However, despite listed efforts, the state of zero-shot cross-lingual generation still remains unclear and poses open questions: %First, 
\begin{itemize}[itemsep=0.cm,topsep=0.1cm,leftmargin=*]
%[noitemsep,parsep=0pt,partopsep=0pt]
\item \textit{Which adaptation method is most effective? }
%to use in practice?}
Methods proposed for mitigating generation in the wrong language, were all tested on different tasks and benchmarks, and not compared to methods from other works, making it hard to establish the best performing one. 
\item \textit{What makes a better mPLM for zero-shot cross-lingual transfer?}
Different models have different pretraining objectives, training and architectural choices. How do these factors impact the quality of the cross-lingual transfer in generation?
\item \textit{Importance of hyperparameters in downstream task adaptation.}
None of the previous work studied the impact of hyper-parameters used during downstream task adaptation for zero-shot cross-lingual transfer in generation.
\item \textit{Finally, if we pick the best solutions from all of the three listed dimensions, how far in performance can we get?}
Can we reach the performance of a strong baseline, data translation, consisting in translating the train data into target languages?
Previous studies either did not reach its performance or did not compare to this baseline.
\end{itemize}

The contribution of this work is conducting a deep empirical study addressing the listed questions. We  consider most commonly used multilingual encoder-decoder mPLMs, namely mT5 and mBART, as well as the translation model NLLB-200. We systematically study six adaptation methods, investigate the effect of intermediate tuning, pay attention to adaptation hyperparameters, and compare models and adaptation methods \textit{in a unified setting}. We consider two tasks: summarization and questions answering (QA). Our main findings are as follows:
\begin{itemize}[itemsep=0.cm,topsep=0.1cm,leftmargin=*]
    \item Hyperparameter tuning plays a very important role in cross-lingual transfer in generation: while related works report a severe problem of generation in a wrong language after full finetuning, we find that simply reducing the learning rate helps to alleviate this problem almost completely, without hurting performance;
    \item Intermediate tuning substantially improves performance in the majority of cases;
    \item With carefully chosen learning rates and intermediate tuning when necessary, simple full finetuning is a very strong baseline for zero-shot cross-lingual transfer in generation. Improvements brought by more advanced methods are quite modest, and none of the methods consistently outperform full finetuning in all cases. The notable methods are freezing the model decoder and embeddings, which performs consistently well with mBART (but not with mT5), and using more than one source language, which performs consistently well with mT5 (but not with mBART);
    \item mBART and mT5 of similar sizes lead to comparable performance. 
    Qualitatively, due to the specifics of the masking pretraining objective, mBART is better suited for tasks with long outputs while mT5 is for tasks with short outputs;
    \item 
    NLLB-200 is surprisingly competitive in summarization, reaching performance of mT5 and mBART  for high-resource Latin-alphabet languages, but lags behind in QA;
    \item 
    The final performance of the zero-shot approach is the same or superior to the performance of the data translation approach, often considered as an upper baseline for cross-lingual transfer in generation.
    Notably, careful learning rate tuning coupled with intermediate tuning allows the zero-shot approach closely approach the performance of data translation simply with the full finetuning adaptation. 
\end{itemize}

\section{Related Work}
All works on zero-shot cross-lingual transfer in generation underline (and try to address) the severe problem of generating in a wrong language at the test time. This problem is also referred to under terms catastrophic forgetting (of languages not participating in finetuning, \citealp{vu-etal-2022-overcoming}), source language hallucination~\citep{mmt5}, or accidential translation problem~\citep{li-murray-2023-zero}. \citet{vu-etal-2022-overcoming} propose to overcome generation in a wrong language by using parameter-efficient finetuning  instantiated by prompt-tuning~\citep{lester-etal-2021-power}. They also  mix-in the unsupervised target language task together with the supervised source language task, and factorize learnable prompts into language and task components.

\citet{mmt5} propose mmT5 (modular mT5), allocating a small amount of language-specific parameters in the model during pretraining and freezing them during task-specific finetuning. To alleviate generation in a wrong language, they freeze some additional mmT5 parameters during finetuning, e.~g.~the embedding layer and feed forward layers in Transformer decoder. \citet{li-murray-2023-zero} argue that learning language-invariant representations during finetuning is harmful for cross-lingual generation and propose finetuning on the data from more than one source language to avoid generation in a wrong language, with mT5 as a backbone model. ZMBART~\citep{zmbart} is the only work which considers the other backbone model than mT5: they perform an intermediate tuning of mBART on an auxiliary unsupervised task on Hindi, Japanese and English. To avoid generation in a wrong language, they freeze embeddings and the Transformer decoder, and mix-in the data from auxiliary pretraining during finetuning.

In our work we are interested to compare all previously proposed approaches in a unified setting, to better assess the impact of different factors on the zero-shot cross-lingual transfer in generation.

Alternative approaches to zero-shot cross-lingual transfer include data translation approaches, often referred as \textit{translate-train} and \textit{translate-test} paradigms. The former one implies translating the train task data to target languages and finetuning the model on this translated data, and the latter one implies translating test input examples into the source language, generating outputs in the source language and translating them back into the target language. The drawbacks of these approaches include a high computational cost either at training or testing time, lack of high-quality translation models for low-resource languages, and potential inconsistencies between sentences in translation~\citep{vu-etal-2022-overcoming}. Despite its computational cost,
data translation is a strong approach which is usually considered as an upper baseline for zero-shot approaches.
Another related field is \textit{few-shot cross-lingual transfer in generation} which assumes access to a small amount of labeled examples in the target language~\cite{schmidt-etal-2022-dont,lauscher-etal-2020-zero, zhao-etal-2021-closer}. This setting is out of scope of this study, but could be considered in the future work.

\section{Methodology and experimental setup}
\label{sec:setting}
\paragraph{Adaptation methods.}
We investigate the following adaptation methods:
\begin{itemize}[itemsep=0.cm,topsep=0.1cm,leftmargin=*]
    \item \textit{Full finetuning}: all weights of the model are finetuned on the source language data;
    \item \textit{Prompt tuning}~\citep{vu-etal-2022-overcoming}: comprises prepending several learnable vectors ("prompt") to the list of embeddings of the text input and freezing all other model weights during finetuning. Parameter-efficient approaches were shown in the literature to be better suited for transfer learning than full finetuning. %LINKS!
    \item \textit{Adapters}~\citep{houlsby, bapna-firat-2019-simple}: lightweight tuned
modules inserted after each fully-connected and attention block of Transformer, when the rest of (pretrained) model weights are frozen. We consider adapters as the most widely used parameter-efficient adaptation approach in the literature;
    \item \textit{Freezing of the decoder and embeddings}~\citep{zmbart}: only weights in the encoder are finetuned. The motivation behind this approach is that the decoder should preserve capabilities of generating in various languages while the encoder will adapt the model to the task;
    \item \textit{Mixing-in self-supervised data for target languages}~\citep{lester-etal-2021-power,zmbart}: during finetuning, task data instances in the source language will be alternated with self-supervised data instances in target languages. The motivation is that such a mixing will preserve model's capability of generation in target languages;
    \item \textit{Using several source languages}~\citep{li-murray-2023-zero}: performing finetuning on more than one source language to better decouple task knowledge from language knowledge.  
\end{itemize}
In the rest of the text term "full finetuning" refers to the finetuning of all weights on the English task data only, even though two last described methods also finetune all weights. We do not consider mmT5 as it was not publicly released and requires substantial resources for pretraining. 

We also experiment with \textit{intermediate tuning} (IT) of the model, used in several works and performed before finetuning on the task data. 
Standard encoder-decoder mPLMs rely on a self-supervised denoising training, where often the input corresponds to corrupted text (eg. with masked tokens or permuted sentences), and the output can follow some very specific structure (eg. a masked span rather than a full sentence, output containing special tokens, etc.). Therefore, in their raw form, these mPLMs are not necessarily well suited to receive well-formed text as an input and generate clean text as an output.  IT performs finetuning on a language modeling-like task, e.g. predicting the continuation of a paragraph based on its beginning, to compensate for this gap.
IT was shown to be necessary in \citet{vu-etal-2022-overcoming} for prompt tuning of mT5 and in \citet{zmbart} for full or partial finetuning of mBART. We systematically test the necessity of IT for all methods and models.

\paragraph{Models.}
We focus on encoder-decoder mPLMs as they are well suited for generation purposes, as opposed to encoder-only mPLMs such as mBERT or XLM-R. We leave the investigation of decoder-only mPLMs such as BLOOM~\citep{bloom} for future work. We consider mT5 and mBART as two most widely used mPLMs and NLLB-200 as a high-quality translation model:
\begin{itemize}[itemsep=0.cm,topsep=0.1cm]
    \item \textit{mT5}: pretrained using the masked language modeling objective where parts of the input sequence are masked and the missing fragments act as targets\footnote{In contrast to English-centric T5, mT5 did not include supervised tasks in pretraining.}. 
    mT5 is pretrained on the mC4 corpora, supports 101 languages, and does not use any language codes. Among released sizes from 300M to 13B we experiment with mT5-base (580M, most of the experiments) and mT5-Large (1.2B, additional experiment).
    \item \textit{mBART (pt)}: pretrained using the denoising objective where parts of the input sequence are masked and the entire original sequence acts as a target~\citep{mbartpt, mbarttr}. mBART is pretrained on Common Crawl~\cite{commoncrawl} corpora, supports 50 languages, has 680M parameters in total and uses language codes in both encoder and decoder sides. Both input sequence \verb|X| and target sequence \verb|Y| are prepended with the language code: \verb|[lang_code, X]| and \verb|[lang_code, Y]|, and at the inference time \verb|lang_code| is forced as a first generated token. Our hypothesis is that the use of the language code in the decoder can help to alleviate the problem of generation in a wrong language.
    \item \textit{mBART (tr)}: In addition to the \textit{pretrained} version, we also consider mBART finetuned for \textit{translation}~\citep{mbarttr}.
    \item \textit{NLLB-200}: translation model supporting 200 languages, pretrained on sentence-level data mined from the web and automatically paired using multilingual embeddings. NLLB-200 uses the same language code scheme as mBART and is released in various sizes from 600M to 54.5B, among them we consider 600M (distilled version). Our hypothesis is that translation-based pretraining may provide good representations for cross-lingual transfer as suggested by \cite{reid-artetxe-2023-role}.
\end{itemize}

\paragraph{Evaluation.}
We select two generative tasks to evaluate cross-lingual zero-shot knowledge transfer: 
\begin{itemize}
    \item \textit{XL-Sum}: news summarization on the XL-Sum dataset~\citep{xlsum}. The model needs to generate a 1--2 sentences summary based on a 1--2 news paragraphs. The evaluation is performed with ROUGE-2 metric~\cite{rouge} computed on the test sets (first 2k examples per language). 
    \item \textit{XQuAD}: question answering dataset ~\citep{xquad}, the model needs to generate a short phrase answer based on a paragraph and a question about it appended in the end of the paragraph. The evaluation is performed with F-measure comparing tokens in the gold answer and the model-generated answer, computed on publicly available development sets. For better metrics interpretability, we only consider questions for which groundtruth answers do not contain numbers and are correctly identified to be written in the target language. 
\end{itemize}
We select a representative subset of languages for each task\footnote{XL-Sum:  Chinese, French, Korean, Russian, and Spanish. XQuAD: Arabic, Chinese, German, Russian, and Spanish}, covering Latin- and non-Latin scripts, and report how do  task-specific metrics evolve during adaptation. For better interpretability, in addition to the task metrics, we also consider  (1) \textit{lang. correct rate} metric (the percentage of outputs generated in the correct target language)  and (2) \textit{average sequence length} metric that allows us to spot some edge behaviour of the models. 
\paragraph{Adaptation settings.}
For all adaptation methods 
we train models on English data for 20k steps with a batch size of 4000 tokens on a single A100 GPU, and run evaluation each 2k steps. We crop input sequences to the maximum length supported by models, which equals to 512 (mT5, NLLB-200) or 1024 tokens (mBART). We grid search the learning rate (LR) for each task-model-adaptation method combination, details are given below.

 For \textit{Intermediate tuning} (IT) we finetune models for 100k steps on the CommonCrawl data uniformly sampled across all target languages and English, with the batch size of 5k tokens and the LR chosen to optimize fluency of model generations, inspected manually. We use  PrefixLM-inspired self-supervision from~\cite{vu-etal-2022-overcoming}, where the continuation of the text needs to be predicted based on its beginning. It has shown more promising results in our preliminary experiments compared to self-supervised objective from~\cite{zmbart} (see details in Appendix~\ref{app:it}).

\begin{itemize}
    \item \textit{Prompt tuning}: we use the prompt dimension of 100 and initialize the prompt with randomly selected rows of the embedding matrix, following \citet{vu-etal-2022-overcoming}.
    \item \textit{Adapters}: we use the adapter dimension of 64 and insert adapters after each attention and fully-connected layer, following \citet{bapna-firat-2019-simple}. 
\item \textit{Mixing-in target languages}: we use the same self-supervised objective as in IT and sample the corresponding data with probability 1\% (all languages represented uniformly within this 1\%), following \citet{vu-etal-2022-overcoming}. We experimented with higher portions in Appendix~\ref{app:mixin}, as well as with mixing-in the pretraining task of the base model, and found that they lead to worse results.
\item \textit{Using several source languages}: we test this approach only on XL-Sum, because for XQuAD only English training data is available; for XL-Sum we use English, Japanese and Arabic, selecting them uniformly when forming mini-batches. 
\end{itemize}

\paragraph{Hyperparameter tuning.}
\label{par:exp_settings}

We tune LR and decide on the necessity of IT, for each considered task-model-adaptation method combination. We initially grid searched LR for full finetuning, adapters and prompt tuning, for each task and model, without IT. 
The result of this step is the preliminary LR (PLR), and we utilize the PLR of full finetuning for other adaptation methods since they are also based on full finetuning. PLR usually corresponds to the highest LR which still enables generation in the correct language. After finding PLR, for each task-model-adaptation method combination, we select the best of four hyperparameter combinations: two options for LR (PLR and PLR $\times 10$) and two options for IT (used or not). Our intuition is that the use of advanced adaptation methodology or IT could potentially increase the LR which still does not lead to generation in the wrong language. In practice, this happened only once, for freezing of mBART in the summarization task. 

For XL-Sum, we perform the described tuning on the validation sets, looking at the performance averaged over considered target languages, while the main evaluation is performed on the test sets. For XQuAD, only validation sets are publicly available so we perform tuning using held-out languages (Thai, Romanian, and Vietnamese). In Appendix~\ref{app:valid} we show that performance on  validation sets in target languages correlates with performance on validation sets in held-out languages and validation sets translated from English into target languages. This demonstrates that having validation sets in target languages is not necessary in practice which is important to enable fully zero-shot setting.
%Results are usually consistent between languages. 

We report the resulting optimal settings in Table~\ref{tab:lrs} in Appendix. We could not find information on the used LR in \citep{mmt5} and \citep{vu-etal-2022-overcoming}, to compare our chosen LRs with theirs. \citet{zmbart} and \citet{li-murray-2023-zero} use a constant LR for all tasks, which are hard to compare to ours because of different data\footnote{\citet{zmbart} use LR=3e-5 larger than ours 1e-6, \citet{li-murray-2023-zero} use LR=7e-5 close to ours 1e-4.}.

More details on the experimental setting are given in Appendix~\ref{app:exp_details}.

\section{Experiments}
\label{sec:exps}

\begin{figure*}[t!]
    \centering
         \includegraphics[height=3.15cm]{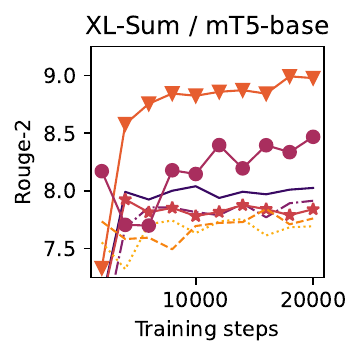} 
          \includegraphics[height=3.15cm]{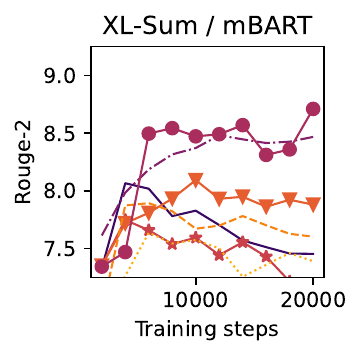} 
         \includegraphics[height=3.15cm]{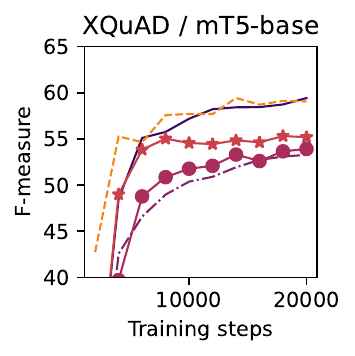} 
          \includegraphics[height=3.1cm]{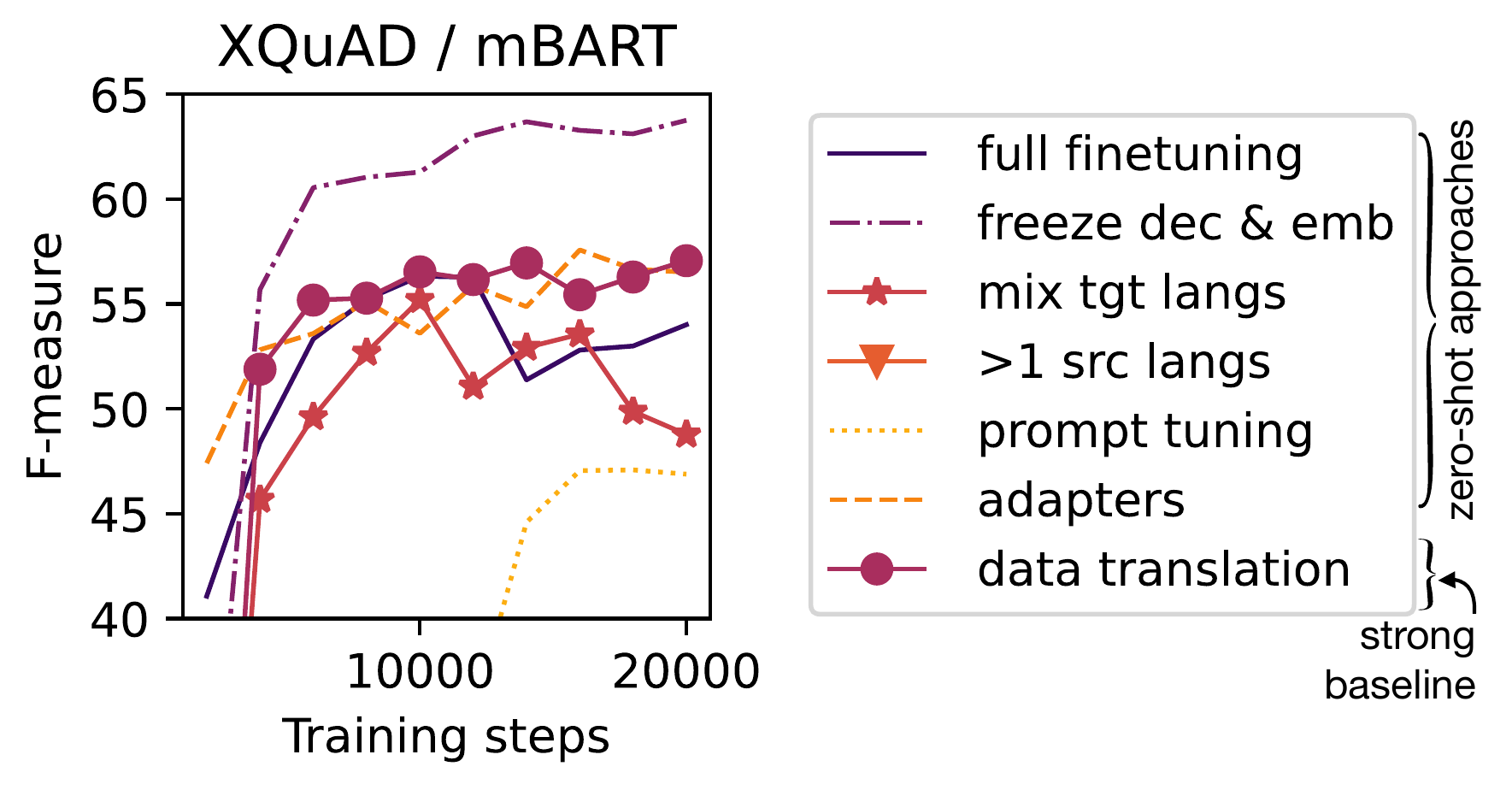} 
        \caption{Comparison of adaptation methods, with tuned learning rates and intermediate tuning when it is needed. 
        Results averaged across target languages and 2 runs. Language correct rate is close to 100\% in almost all cases, due to hyperparameter tuning. The exception is prompt tuning of mT5 in the XQuAD task which is not shown because of too low performance. \textit{Main conclusions}: (1) Straightforward full finetuning is a strong approach which reaches or approaches the performance of data translation in all cases. (2) None of other approaches outperform full finetuning \textit{consistently} in all cases: using several sources languages works well for mT5 but not for mBART and freezing decoder works well for mBART but not mT5. (3) One of zero-shot approaches reaches or outperforms a strong and computationally expensive baseline, data translation, in all cases.}
        \label{fig:methods}
\end{figure*}

First, we investigate the effect of the learning rate, intermediate tuning and the adaptation method for two most commonly used models, mT5 and mBART. Second, we compare them with other models and consider larger models. Finally,  we present some qualitative examples and observations from manual inspection of predictions. In general, model predictions reaching highest metric values in our plots, form quite meaningful and reasonable responses to the considered tasks; more details in Section~\ref{sec:inspection}.

\paragraph{Effect of learning rate.} 
We begin our study with analysing the effect of LR on the full finetuning on the English task data.
With too small or too large LR the model does not learn even the English task because of too short steps or divergence. 
For the range of LRs when the English task is learned well, we observe that larger LRs lead to the effect reported in other works, when the model overfits to the source English language and generates answers in English when applied for inputs in other languages. However, \textit{with the reduced LR, this effect almost completely eliminates and the model mostly generates in the target language}. 
This effect is demonstrated in Figure~\ref{fig:lr_ft} on a subset of languages and in Fig.~\ref{fig:mt5_xlsum}--\ref{fig:mbart_pt_xquad} in Appendix on all considered languages.

Figure~\ref{fig:lr_ft} also shows a comparison of enhancements of full finetuning proposed in the literature, such as mixing-in target languages or freezing the decoder and embeddings. Even though these enhancements improve performance and percentage of outputs in the correct language with fixed LR, we find that \textit{reducing LR in full finetuning often brings larger improvements}. Reducing LR for other methods makes them even stronger. 

We note that performance in English is usually a little higher with larger LR. This may raise a hypothesis that for non-English languages, outputs generated with larger LR in English may be of higher semantic quality than the ones generated in the correct target language with smaller LR. In Appendix~\ref{app:addexp} we test this hypothesis and demonstrate that this is not the case.

\begin{table}[t]
\centering
\begin{small}
\begin{tabular}{p{2.2cm}|cccc}
 \toprule
& \multicolumn{2}{c}{XL-Sum} &  \multicolumn{2}{c}{XQuAD} \\
\midrule 
Method & mT5 & mBART & mT5 & mBART   \\
\midrule 
Full finetuning    &  +0.1   &  +2.5   &   +6.3  &  +9.0      \\
Ft + mix tgt langs    &  0   & +0.6   &  +3.1   &   -8.3     \\
Ft + >1 src langs    &   0  &  +1   &  n/a   &   n/a     \\
Freeze emb\,\&\,dec    &   +4.3  &  +4.1  &  +11.2    &   +1.3     \\
Adapters    &  0   &  0   &   +1.0  &   +3.9     \\
Prompt tuning    &  +7.5   &  +7.2   &   +26.8  &     +25.1   \\
\bottomrule
\end{tabular}
\end{small}
\caption{\label{tab:it}
Difference in performance between task adaptation with and without intermediate tuning,
for various methods. Rouge-2 for XL-Sum, F-measure for XQuAD. 
\textit{Main conclusion:} intermediate tuning brings performance improvements in the majority of cases, in almost all the rest cases it does not affect performance.
}
\end{table}

\paragraph{Effect of intermediate tuning.}  
For each combination of a task and an adaptation method, we compare the  mT5-base/mBART task adaptation with and without intermediate tuning (IT). 

We choose the best LR between PLR and PLR $\times 10$ (section ~\ref{par:exp_settings}). Results are presented in Table~\ref{tab:it}. 
We observe that \textit{intermediate tuning substantially increases performance in the majority of cases}. 
In particular, IT appears to be essential for mBART with almost all adaptation methods and in all tasks, and important for mT5 in question answering. For mT5 in summarization, the use of IT does not increase performance, except with prompt tuning and freezing methods. We believe  this is because  these two approaches do not modify the decoder, which was trained only on masked spans targets during mT5 pretraining and was never exposed to realistic text targets, 
and IT closes this gap. This result is consistent with \cite{vu-etal-2022-overcoming} and \cite{zmbart}.

\paragraph{Comparison of adaptation methods.} 
Figure~\ref{fig:methods} shows results (averaged over target languages) comparing adaptation methods for mT5-base and mBART models. 
Detailed per-language results are presented in Figure~\ref{fig:methods_langs} in Appendix.

We observe that \textit{with carefully chosen learning rates and intermediate tuning, simple full finetuning is a very strong baseline for zero-shot cross-lingual transfer in generation}. Improvements brought by the use of more advanced adaptation methods are rather modest, and \textit{none of the adaptation methods consistently outperform full finetuning in all cases}. \textit{The notable approach for mBART is freezing the decoder and embeddings}, proposed by~\citet{zmbart} for this base model: freezing consistently outperforms full finetuning in all target languages in both tasks. However, this approach does not show such improvements for mT5. \textit{For XL-Sum, using more than one source language proposed in~\cite{li-murray-2023-zero} for mT5, brings consistent improvement over target languages} when used with mT5. For mBART this approach performs on par with using one source language. 
The obvious drawback of this approach is that multi-lingual data may  be not available, e.g. this is the case for XQuAD.

Mixing-in unsupervised tasks for target languages often degrades performance and increases the length of predictions, see Appendix~\ref{app:mixin}. Prompt tuning often has difficulties learning an English task and substantially underperforms other adaptation methods on XQuAD. Adapters usually perform on par or slightly worse than full finetuning.

\begin{figure*}[t!]
    \centering
         \includegraphics[width=\linewidth]{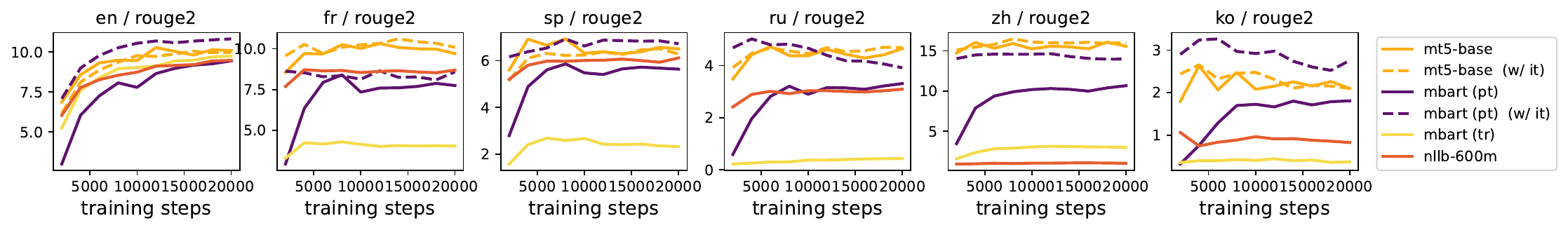} 
        \includegraphics[width=\linewidth]{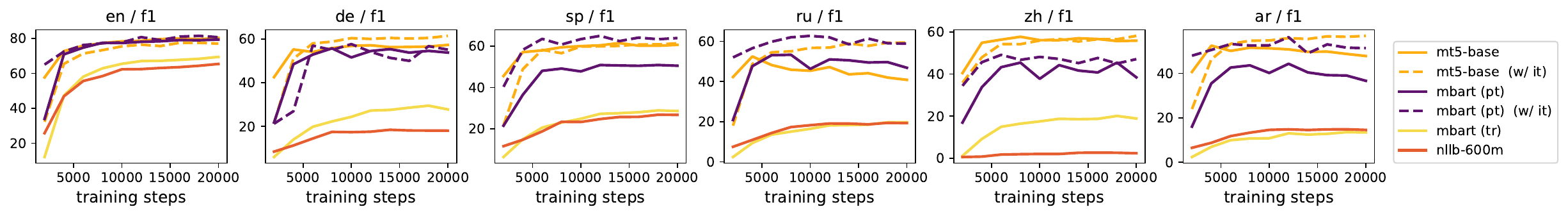} 
        \caption{Comparison of base models with full finetuning. Each plot averaged over 3 runs. Correct language rate is close to 100\%, due to hyperparameter tuning, in almost all cases except the translation-tuned version of mBART. pt: pretrained version of mBART, tr: translation-finetuned version of mBART. \textit{Main conclusion:} mBART and mT5 of similar sizes perform on par; NLLB performs well in summarization for Latin-alphabet languages.}
        \label{fig:models}
\end{figure*}

\begin{figure*}[t!]
    \centering
         \includegraphics[width=\linewidth]{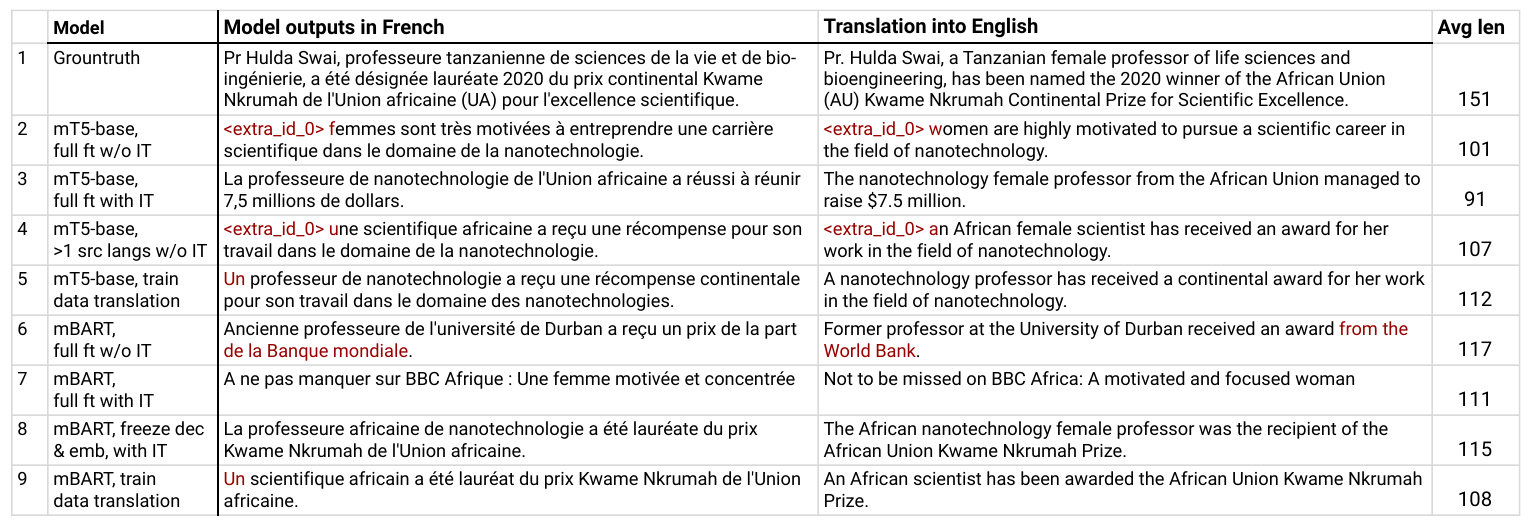} 
        \caption{Example predictions for a selection of models. Avg. len. over evaluation corpora in French, in characters. Red highlights errors or extra tokens.}
        \label{fig:examples}
\end{figure*}

\paragraph{Comparison of models.}
Figure~\ref{fig:methods} allows us to compare mT5-base and mBART after tuning of hyperparameters and adaptation methods. These models incorporate comparable numbers of parameters. We observe that \textit{mT5 and mBART reach the close level of performance in both tasks}. The same conclusion holds if we simply compare full finetuning runs of both models. 

\begin{table}[t]
\centering
\begin{small}
\begin{tabular}{p{2.9cm}|cccc}
 \toprule
& \multicolumn{2}{c}{XL-Sum} &  \multicolumn{2}{c}{XQuAD} \\
\midrule 
Method & R2 & LCR & F1 & LCR   \\
\midrule 
Large / IT + ft & 9.9 & 99.8\% & 69.8 & 94.7\%\\
Large / IT + ft >1 src lg &  10.9 &  99.8\% & n/a  & n/a \\
Large / Data translation & 10.8 & 99.8\% & 63.6 & 96.7\%\\ 
\midrule
Base / IT + ft & 8.0 & 99.7\% & 59.4 & 92.9\%\\
Base / IT + ft >1 src lg &  9.0  & 99.8\%  &  n/a & n/a \\
Base / Data translation & 8.5 & 99.6\% & 53.9 & 95.3\%\\
\bottomrule
\end{tabular}
\end{small}
\caption{\label{tab:large}
Results for mT5-large model, averaged over target languages. Metrics: Rouge-2 for XL-Sum, F-measure for XQuAD, LCR: language correct rate. LCR is lower than 100\% on XQuAD (partly) because of language identification errors for short sequences. 
}
\end{table}

In Figure~\ref{fig:models} we compare all four models we consider, adapted using full finetuning. 
We compare models without intermediate tuning, to avoid hindering model capabilities behind this additional step. We find that \textit{translation-pretrained NLLB-200 performs well in summarization}, achieving performance of mT5 and mBART in Latin-language high-resource languages, French and Spanish, and performing on par with mBART without intermediate tuning in other languages\footnote{Expect Chinese, for which NLLB-200 generates a lot of empty predictions. NLLB-200 was noticed previously in the literature to have issues with processing Chinese.}. We selectively inspected the predictions of NLLB and found that they indeed form meaningful summaries. However, in QA, NLLB-200 performs poorly, often (but not always) generating non-relevant answers. Translation-finetuned version of mBART performs poorly in all tasks, generating a lot of wrong language predictions. 

Interestingly, the results do not support our initial hypothesis that the translation pretraining objective used in NLLB and architectural choices such as the use of language codes in mBART, could improve zero-shot knowledge transfer in generation.

\paragraph{Comparison versus data translation.} 
Figure~\ref{fig:methods} also shows comparison versus the data translation\footnote{Data translation is often referred as translate-train method.} approach, when English training data is translated into target languages using the NLLB-3.3B model. We translate data sentence-by-sentence and grid search the LR for finetuning. The results show that after careful tuning, \textit{the zero-shot approach reaches or outperforms the data translation approach in both considered tasks}. If we consider a simpler setting when only LR and the use of IT are tuned, i.e. comparing full finetuning and data translation runs in Figure~\ref{fig:methods}, we observe that the zero-shot approach closely approaches the data translation approach in summarization and performs on par in question answering. The XQuAD dataset is harder to automatically translate than XL-Sum, 
e.g. single words often present in targets may be translated into short full sentences.

\paragraph{Experiments with larger models.}

Table~\ref{tab:large} reports results for the mT5-large model where we compare performance achieved with full finetuning after intermediate tuning versus training on translated data. We also include the leader approach of using several source languages for XL-Sum. We consider only mT5 because mBART is released in one size.  We reduce LR to 0.00001 for the larger model, as the LR of 0.0001 used for the base model was sometimes producing English outputs. We also list mT5-base results for reference.

We find that the same conclusions hold for the mT5-large model as for mT5-base: reducing LR eliminates generation in the wrong language, and  the zero-shot approach is on par or better than the data translation approach.

\section{Inspection of predictions} 
\label{sec:inspection}
We inspected a subset of predictions in French and Russian and found that models achieving highest scores in both tasks generate fluent, meaningful and reasonable predictions in a lot of cases, but sometimes have issues with factualness, grammaticality or hallucinations. Examples are shown in Figure~\ref{fig:examples}. Analyzing effects of LR, we observe that increasing LR leads first to increase in code switching and then to wrong language generation, while \textit{reducing LR leads to producing rudiments of pretraining in generation}. For example, models sometimes generate extra tokens used in pretraining, such as \verb|<extra_id_{N}>| for mT5 or \verb|<sep>| for mBART, see rows 2 and 4 in Figure~\ref{fig:examples}. \textit{In most cases this does not affect meaningfulness of predictions}, but in \textit{rare} cases leads to mT5 producing incomplete sentences, which may look unreasonable in summarization, e.g. ``\texttt{<extra\_id\_0> Guinea-Bissau President Alberto Dabo said.}'' (translated from French). The reason is that in mT5 pretraining tokens \verb|<extra_id_{N}>| were followed by fragments of input sentences. \textit{The described effect is eliminated by intermediate tuning} (row 3 in Fig.~\ref{fig:examples}).

In the same fashion, \textit{mBART average lengths are closer to groundtruth average lengths than mT5 in summarization, and the reverse effect takes place in QA}. The reason is that in mT5 pretraining, the targets are only fragments masked in the input, which are shorter than targets in mBART pretraining represented by full sequences (they need to be reconstructed from the masked inputs).

Notably, data translation can  produce translation-related errors, e.g. in rows 5 and 9 models generate a wrong male article "Un", probably because this was a dominating article in the translated data.

\section{Conclusion}
In this work, we conducted a deep systematic study of how to achieve high-performing zero-shot cross-lingual transfer in generation. Our study highlights the \textit{high importance of careful learning rate tuning and the usefilness of the intermediate tuning.} We show that with these two ingredients, mT5 and mBART achieve strong results with simple full finetuning, i.e. closely approach the performance of  translate-train in summarization and reach it in question answering. The performance gap in summarization is closed by using several source languages for mT5 and freezing decoder and embeddings for mBART. 
Translation-pretrained NLLB-200 shows surprisingly good performance in summarization but lags behind in question answering. We urge future works to pay more attention to hyperparameter tuning and to report more rigorously their experimental setup, as well as consider a wider spectrum of models and baselines in the experiments.

\section{Limitations and broader impact}
We aim at conducting a deep, thoughtful study of various design choices in zero-shot cross-lingual generation, but acknowledge the impossibility of considering all possible options, given the resource constraints. In particular, we could not perform full fine-grained grid search of LR for each task-model-adaptation method combination. Instead, we use a well-designed simplified strategy described in Section~\ref{sec:setting}, which already gave strong results. In the same fashion, we had to limit our study to three models (we picked most commonly used models) and adaptation methods which do not require model pretraining, e.g. we do not consider mmT5 model. Nonetheless, we hope our study provides helpful insights on zero-shot cross-lingual transfer in generative tasks and shows that it can achieve the performance of the data translation method, which is usually considered as an unreachable upper baseline.

We do not anticipate any negative impact of our work and on the reverse hope that it will help to make higher-quality language technologies accessible to a broader set of languages.

\section{Acknowledgments}
We gratefully appreciate Alexandre Bérard's and  Sheng Liang's help in the project.

\bibliography{anthology,custom}

\appendix
\clearpage
\section{Experimental setup}
\label{app:exp_details}

\paragraph{Data.} 
We experiment with news summarization on the XL-Sum dataset~\citep{xlsum} (released under the CC BY-NC-SA 4.0 license) and question answering on the XQuAD dataset~\citep{xquad} (released under the CC BY-SA 4.0 license). Both datasets were released for research puposes. The XL-Sum dataset was obtained by crawling BBC news in 44 languages, with corpus size per language varying from 1K (Scottish Gaelic) to 300K (English) article-summary pairs. Inputs are composed of 1--2 paragraphs and targets are usually 2--3 sentences. We evaluate on test sets and crop test sets larger than 2K samples, to 2K. The XQuAD dataset was obtained by translating SQuAD validation set~\cite{squad} into 11 languages, thus all language corpora are parallel. We use this dataset for evaluation and train on the training set of SQuAD (80K training instances). Each input is composed of a paragraph and a question about this paragraph appended in the end of the paragraph. Each output is an answer to a question, a short segment copied from the paragraph.

\paragraph{Preprocessing and postprocessing.} We tokenize data using each model's tokenizer. We crop model inputs and outputs to the maximum lengths supported by models, which equal to 1024 tokens for mBART and 512 tokens for mT5-base and NLLB-600M. Due to the design of pretraining, models may generate extra tokens such as \verb|<extra_id_{N}>| for or \verb|<sep>| for mBART. We remove such extra tokens from predictions before computing metrics.

\paragraph{Models and training.} We consider three models: mT5 (base and large, released under the Apache License 2.0 license), mBART (MIT license), and NLLB-200 (cc-by-nc-4.0 license). All models allow use for research purposes. We train models on English data for 20k steps with batch size of 4000 tokens on a single A100 GPU, and conduct validation on considered target languages each 2k steps. We use Adam optimizer with standard inverse square root LR schedule and warm up of 4k steps, and update model parameters after each mini-batch. We estimated the total computational budget of our experiments to be 4K GPU hours.

\paragraph{Hyperparameter search.} For full finetuning, adapters and prompt tuning, we run a search over a range of LR. For each task and model (without intermediate tuning), we search the LR best for non English languages on average, looking at ROUGE-2 for summarization and F-measure for QA. We start with the set of three LRs: $\{10^{-k}, k=3, 4, 5\}$. If the optimal $k^* \ne 4$ then we extend search correspondingly to $k=2, 1$ or $k=6, 7$  until performance stops improving. For full finetuning, after we find optimal $k^*$ we also consider $3\cdot10^{-k^*}$. The motivation is that the optimal $k^*$ usually corresponds to the maximal $k$ that still allows generation in the correct language, and considering $3\cdot10^{-k^*}$ enables more accurate search for this maximum. We report chosen LRs for full finetuning and adapters in Table~\ref{tab:lrs}. For prompt tuning we chose LR of 0.01 for both tasks.

\begin{table}[]
\centering
\begin{small}
\begin{tabular}{p{1cm}p{2cm}|ll|ll} % p{0.45cm}p{0.45cm}p{0.45cm}p{0.45cm}
\toprule
\textbf{Model} & \textbf{Method} &  \multicolumn{2}{c|}{\textbf{XL-Sum}} &    \multicolumn{2}{c}{\textbf{XQuAD}} \\
& & \textbf{LR} & \textbf{IT?} & \textbf{LR} & \textbf{IT?} \\
\midrule
{} & Ft w/o IT & 1e-4  &  & 1e-4  &  \\
{} & Ft &  1e-4 &  & 1e-4  &$\checkmark$ \\
mT5  & + Mix tgt langs & 1e-4  &   &  1e-4 & $\checkmark$\\
(base) & + >1 src langs &  1e-4 &  & n/a  & \\
 & Freeze &  1e-4 & $\checkmark$ &  1e-4  & $\checkmark$\\
 & Adapters & 1e-3 &   & 1e-3  &\\
 & Prompt tuning & 1e-2 & $\checkmark$  &  1e-2 & $\checkmark$\\
\midrule
{} & Ft w/o IT & 1e-6  &   &  1e-5  & \\
{} & Ft & 1e-6  & $\checkmark$ &  1e-5  & $\checkmark$\\
{} & + Mix tgt langs &  1e-6 & $\checkmark$  &  1e-5  & \\
mBART & + >1 src langs &  1e-6 & $\checkmark$ & n/a &  \\
{} & Freeze & 1e-5  & $\checkmark$  &  1e-5  & $\checkmark$\\
 & Adapters & 1e-5 & $\checkmark$ &  1e-3  & $\checkmark$\\
 & Prompt tuning & 1e-2 & $\checkmark$  &  1e-3   & $\checkmark$\\
 \midrule
 {NLLB} & Ft w/o IT & 1e-5  &   & 3e-5  &\\
% { (600m)} & Ft &   &   &   &\\
 \midrule
 mBART (tr) & Ft w/o IT &  1e-6 &  & 1e-3  & \\
 %(tr) & Ft &   &   &   &\\
\bottomrule
\end{tabular}
\end{small}
    \caption{Best hyperparameter configurations for non-English languages: chosen learning rates and whether intermediate tuning (IT) is used. n/a: not applicable.
    }
    \label{tab:lrs}
\end{table}

\paragraph{Evaluation.} For summarization, we report the ROUGE-2 metric~\cite{rouge}, and for QA, we report F-measure. 
In QA, a lot of answers contain numbers or English words which could inflate metrics even if the model does not generate in the correct language. Moreover, the accuracy of language identification decreases on short answers, resulting in false indication of generation in wrong language. To avoid these issues, we compute metrics in QA only over questions for which groundtruth answers do not contain numbers and are correctly identified to be written in the target language ($\sim$50\% of 1190 questions satisfy this criteria).

For ROUGE metric, we use the \verb|gem-metrics| package. For F1 metric in XQuAD, we use the script provided by the dataset authors. To identify language, we use \verb|fasttext| library~\cite{fasttext1, fasttext2} and its \verb|lid.176.bin| model\footnote{\url{https://fasttext.cc/docs/en/language-identification.html}}.

\section{Preliminary experiments with intermediate tuning}
\label{app:it}

\begin{figure}[t!]
    \centering
    \begin{tabular}{cc}
   \multicolumn{2}{c}{} \\
         \includegraphics[height=2.5cm]{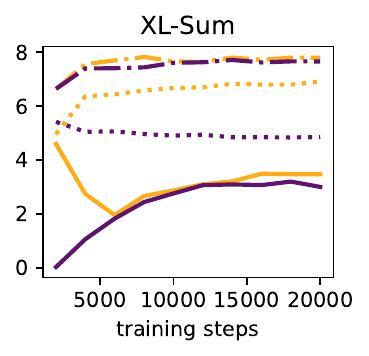} 
        & \includegraphics[height=2.5cm]{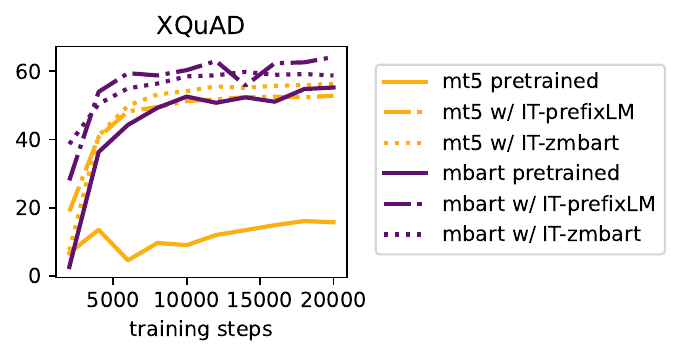} 
        % \\
        % \multicolumn{2}{c}{Correct language rate} \\
        % \includegraphics[height=2.5cm]{figs2/comp_it_xlsum_lid.pdf} &
        % \includegraphics[height=2.5cm]{figs2/comp_it_xquad_lid.pdf} 
        \end{tabular}
        \caption{Comparison of self-supervised objectives for intermediate tuning, with freezing decoder and embeddings as an adaptation method. Task metric: Rouge-2 for XL-Sum, F1 for XQuAD. Correct language rate is close to 100\% in all cases except pretrained mBART on XL-Sum.}
        \label{fig:it}
\end{figure}
Figure~\ref{fig:it} reports comparison of two self-supervised objectives for intermediate tuning: Prefix-LM and ZmBART-like objective. PrefixLM objective implies predicting the continuation of the chuck of text based on its beginning, while ZmBART-like objective implies citing random sentences from the input chunk of text. We compare two objectives using the freezing of the decoder and embeddings as an adaptation method, applied after intermediate tuning with the chosen objective, because we found intermediate tuning to be essential for this adaptation method in the preliminary experiments. Finetuning LR equals to the PLR defined in  Section~\ref{sec:exps}, intermediate tuning LR was chosen to optimize fluency of model generations, inspected manually. Intermediate tuning is performed on the CommonCrawl dataset.

We observe that for XL-Sum, the Prefix-LM objective leads to substantially higher Rouge-2 values, while for XQuAD both objectives lead to close results. Based on these results, we decided to use the Prefix-LM objective in all experiments. 

\section{Preliminary experiments with mixing-in target languages}
\label{app:mixin}

\begin{figure}[t!]
    \centering
    \begin{tabular}{cc}
   \multicolumn{2}{c}{} \\
         \includegraphics[width=0.45\linewidth]{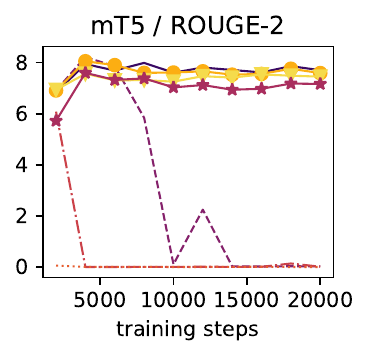} 
        & \includegraphics[width=0.45\linewidth]{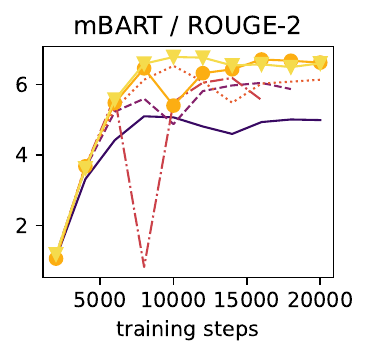} 
         \\
        \includegraphics[width=0.45\linewidth]{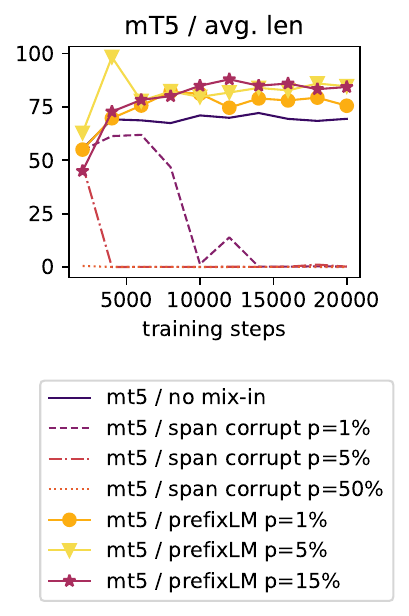} 
        & \includegraphics[width=0.45\linewidth]{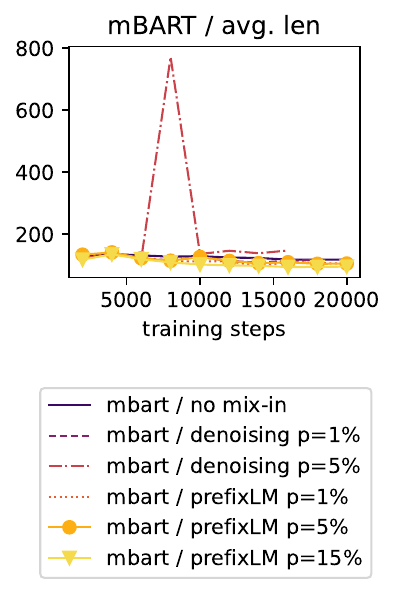} 
        \end{tabular}
        \caption{Preliminary experiments with mixing-in a self-supervised task for target languages. The probability in the legend denotes the probability of sampling target language examples when forming mini-batches. Two self-supervised tasks considered: Prefix-LM and the pretraining task of the model. Correct language rate is close to 100\% in all cases }
        \label{fig:mixin}
\end{figure}

\begin{figure}[h!]
    \centering
        \begin{tabular}{cc}
         \includegraphics[width=0.45\linewidth]{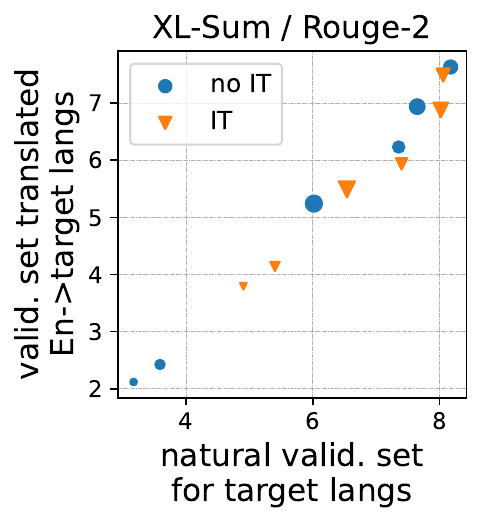} &
         \includegraphics[width=0.45\linewidth]{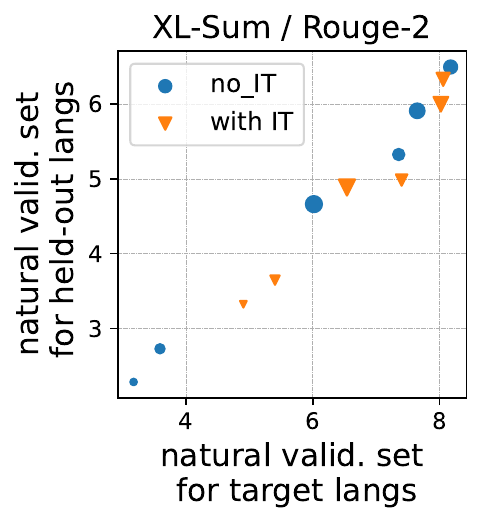} \\
         \includegraphics[width=0.45\linewidth]{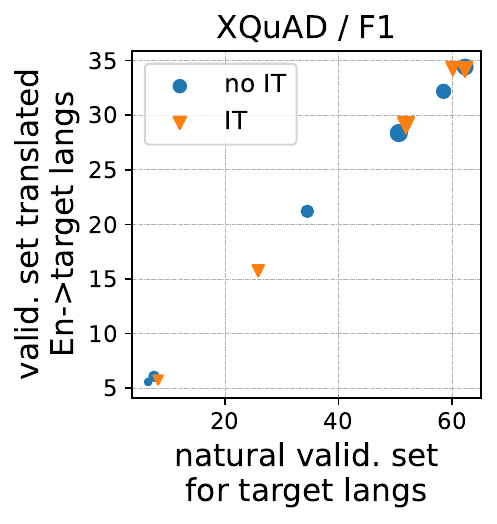} &
         \includegraphics[width=0.45\linewidth]{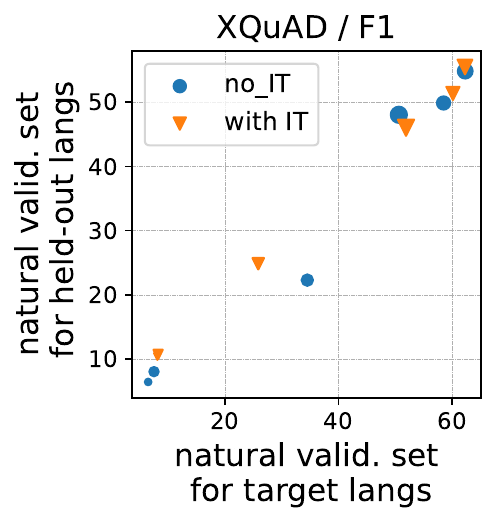}
        \end{tabular}
        \caption{Correlation in performance on various validation sets. Each dot represents mT5-base finetuned on English data, with or without intermediate tuning (reflected with the shape and color of the point), and the size of the point reflects the learning rate. Natural valid. set means the validation set provided by the authors of the dataset.}
        \label{fig:valid}
\end{figure}

Figure~\ref{fig:mixin} reports results of preliminary experiments with mixing-in a self-supervised task in target languages. For each base model, namely mT5-base and mBART, we consider its  pretraining task and a Prefix-LM task used for intermediate tuning.  We consider several options for the probability of sampling target language examples when forming mini-batches. CommonCrawl data is used for the self-supervised task. The experiment is conducted for the XL-Sum task, with LR being equal to the PLR defined in Section~\ref{sec:exps}, without intermediate tuning. 

For mt5, we observe that using the span corruption pretraining task  leads to empty outputs with any mixing-in probability (with smaller probabilities this effect happens later in the training). This is because task examples do not contain any mask tokens, and empty generation is a default response of the pretrained mT5 to such inputs. Mixing-in PrefixLM task examples performs similarly to the standard finetuning of mT5, with mixing-in probability of 1\% performing best, same as in~\citep{vu-etal-2022-overcoming}. Qualitatively, mixing-in self-supervised task increases the length of generated outputs in the tasks of interest.

For mBART, all mixing-in strategies lead to modest improvements in performance, with PrefixLM task performing slightly better. All considered mixing-in probabilities lead to similar results. Based on these observations, we decided to use the PrefixLM task with mixing-in probability of 1\% in our experiments.

\section{Comparing validation sets}
\label{app:valid}

Figure~\ref{fig:valid} demonstrates the correlation between performance measured on various validation sets. Performance on  validation sets in target languages correlates with performance on validation sets in held-out languages and validation sets translated from English into target languages. This shows that having validation sets in target languages is not necessary in practice which is important to enable fully zero-shot setting.

\section{Additional experiment with translating English outputs into target languages}
\label{app:addexp}
\begin{table}[]
\centering
\begin{small}
\begin{tabular}{p{0.3cm}l|cc|cc} % p{0.45cm}p{0.45cm}p{0.45cm}p{0.45cm}
\toprule
{} &{} &  \multicolumn{2}{c|}{\textbf{Best-En LR + Tr.}}  &   \multicolumn{2}{c}{\textbf{Best-non-En LR}}  \\ \midrule
{} &{} &  \textbf{LR} &    \textbf{Score} &    \textbf{LR} &  \textbf{Score} \\
\midrule
{} & mT5 & 1e-3  & 4.02  &  1e-4  & 7.7  \\
Sum & mBART & 1e-5  & 4.06  &  1e-6  & 5.34  \\
{} & NLLB-200 & 1e-4  & 2.86  &  1e-5  & 4.62  \\ 
\midrule
{} & mT5 & 1e-4  &  46.2 &  1e-4  & 58.6  \\
QA & mBART & 1e-5  & 41.1  &  1e-5  & 46.6  \\
{} & NLLB-200 & 1e-4  &  17.4 &  3e-5  &  18.2 \\
\bottomrule
\end{tabular}
\end{small}
    \caption{Comparison of best LR for non-English languages and best LR for English with model outputs being translated into target languages. Performance averaged over non-English languages, after 20k of full finetuning. Reported metric: Rouge-2 for summarization, F-measure for QA. mBART --- pretrained version, no intermediate tuning is used in this experiment.
    }
    \label{tab:translation}
\end{table}

When reducing the LR for preserving generation in correct language, a reasonable question could be whether predictions of higher LR models are higher quality answers, but just in the wrong language, or simply hallucinations caused by data distribution shift. The premise for the former scenario is that on English data, performance with our chosen LR is usually slightly lower than with a larger LR. 

We find that actually the later scenario takes place, by comparing
performance of our chosen LR (best for non-English) and of the best LR for English with model predictions being translated into target languages using NLLB-3.3B\footnote{NLLB-3.3B handles well inputs containing code switching which are frequent in predictions we are translating, and simply copies inputs which are already in the target language.}, for last checkpoints of full models finetuning. According to Table~\ref{tab:translation}, translated predictions of the higher LR model score lower than the (non-translated) predictions of the lower LR model. This result further advocates for the importance of careful LR tuning for full finetuning in zero-shot cross-lingual transfer in generation.

\begin{figure*}[t!]
    \centering
         mT5-base in summarization on XL-Sum \\
         \includegraphics[width=\linewidth]{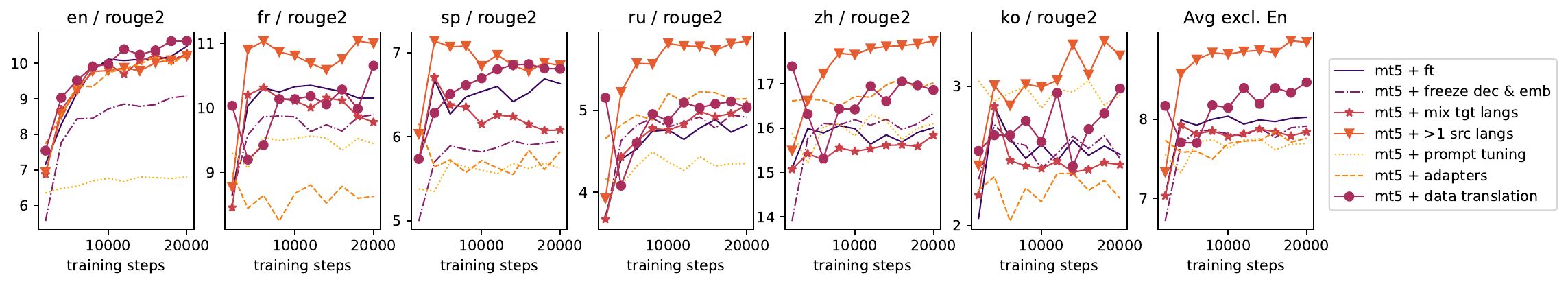}  \\
         mBART  in summarization on XL-Sum \\
         \includegraphics[width=\linewidth]{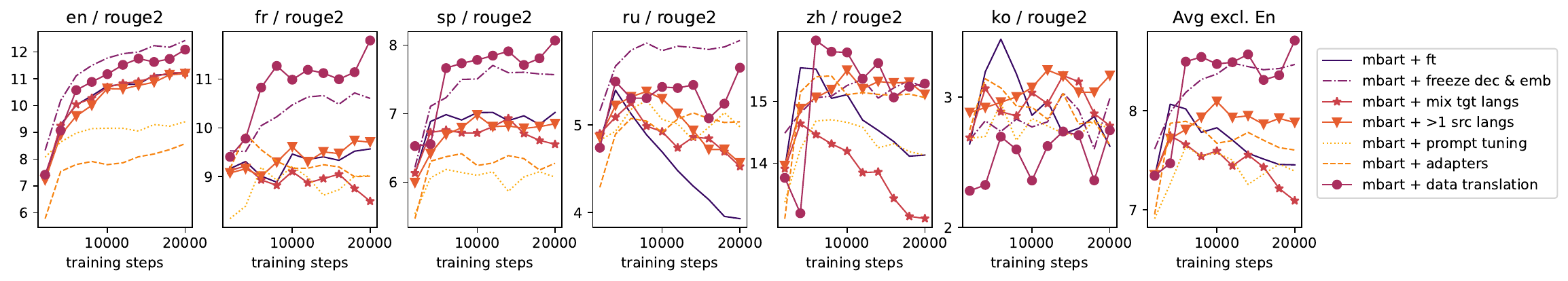}    \\
         mT5-base in question answering on XQuAD \\
         \includegraphics[width=\linewidth]{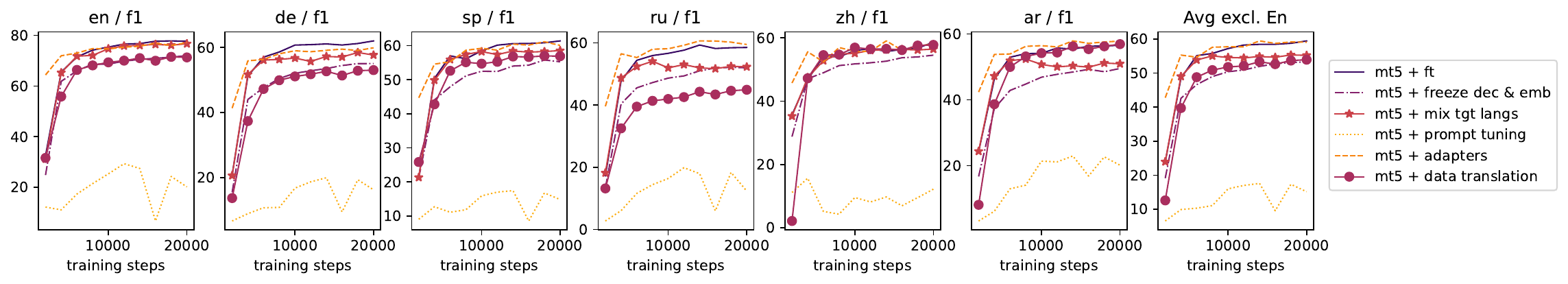} \\
         mBART in question answering on XQuAD
         \includegraphics[width=\linewidth]{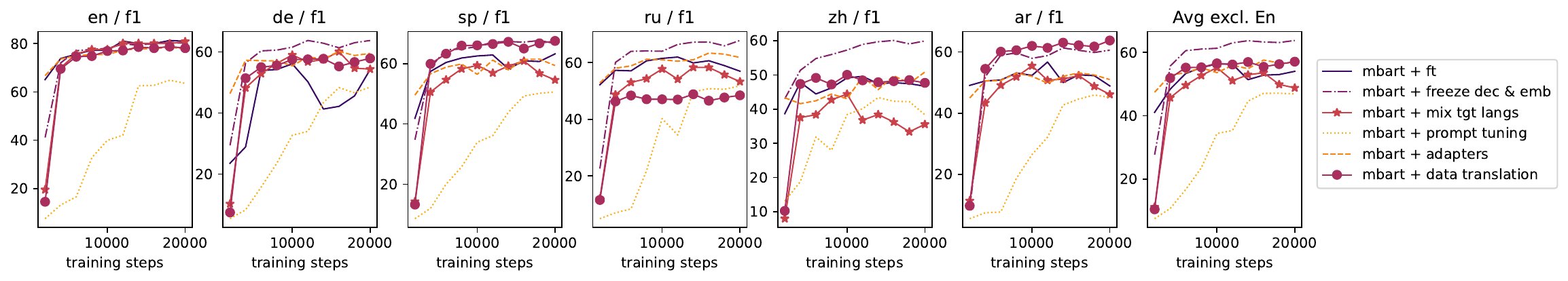}  
        \caption{Per-language results on the comparison of adaptation methods. Each plot averaged over 2 runs. Correct language rate is close to 100\% in all cases, due to the hyperparameter tuning, except prompt tuning of mT5 in the XQuAD task.}
        \label{fig:methods_langs}
\end{figure*}

\begin{figure}
    \centering
     \includegraphics[width=\linewidth]{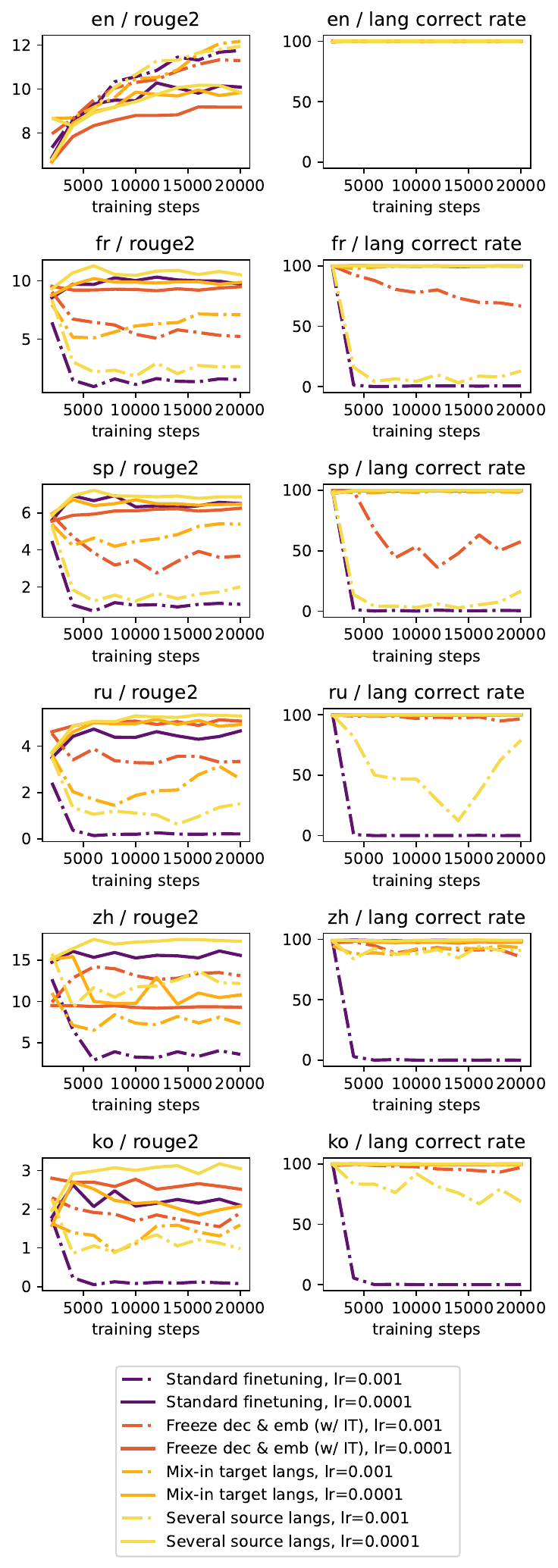} 
    \caption{Per-language results on the effect of learning rate, for mT5 on XL-Sum.}
    \label{fig:mt5_xlsum}
\end{figure}

\begin{figure}
    \centering
     \includegraphics[width=\linewidth]{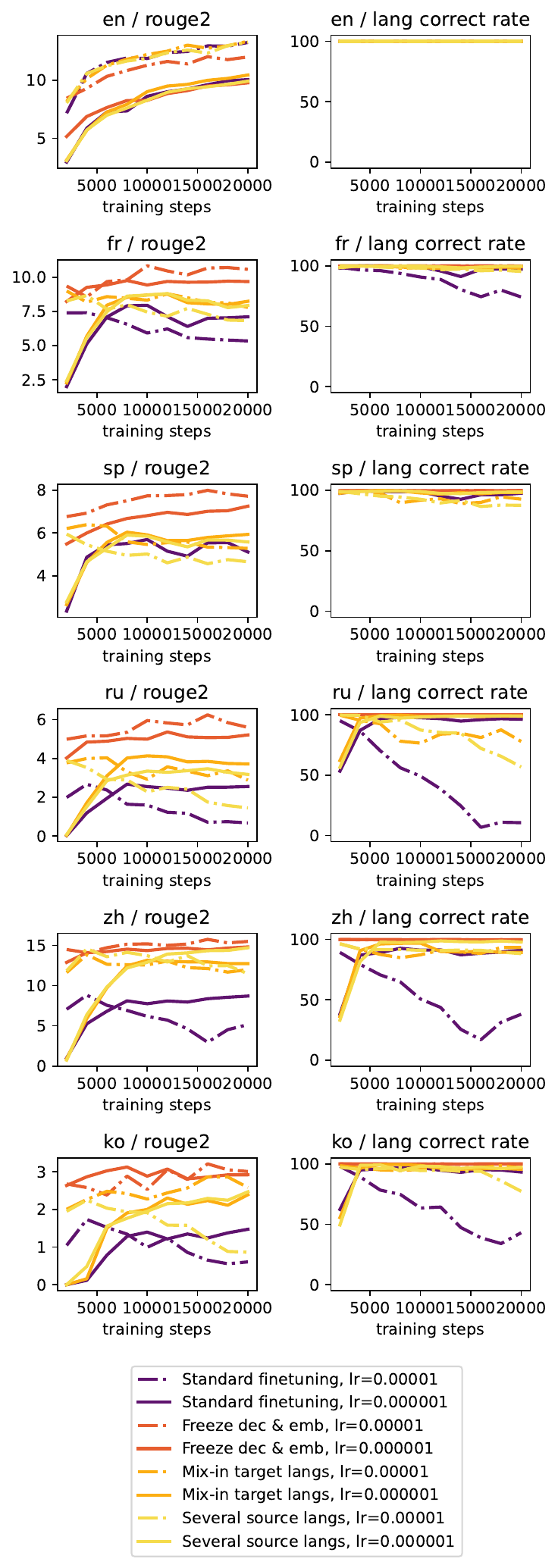} 
    \caption{Per-language results on the effect of learning rate, for mBART on XL-Sum.}
    \label{fig:mbart_pt_xlsum}
\end{figure}

\begin{figure}
    \centering
     \includegraphics[width=\linewidth]{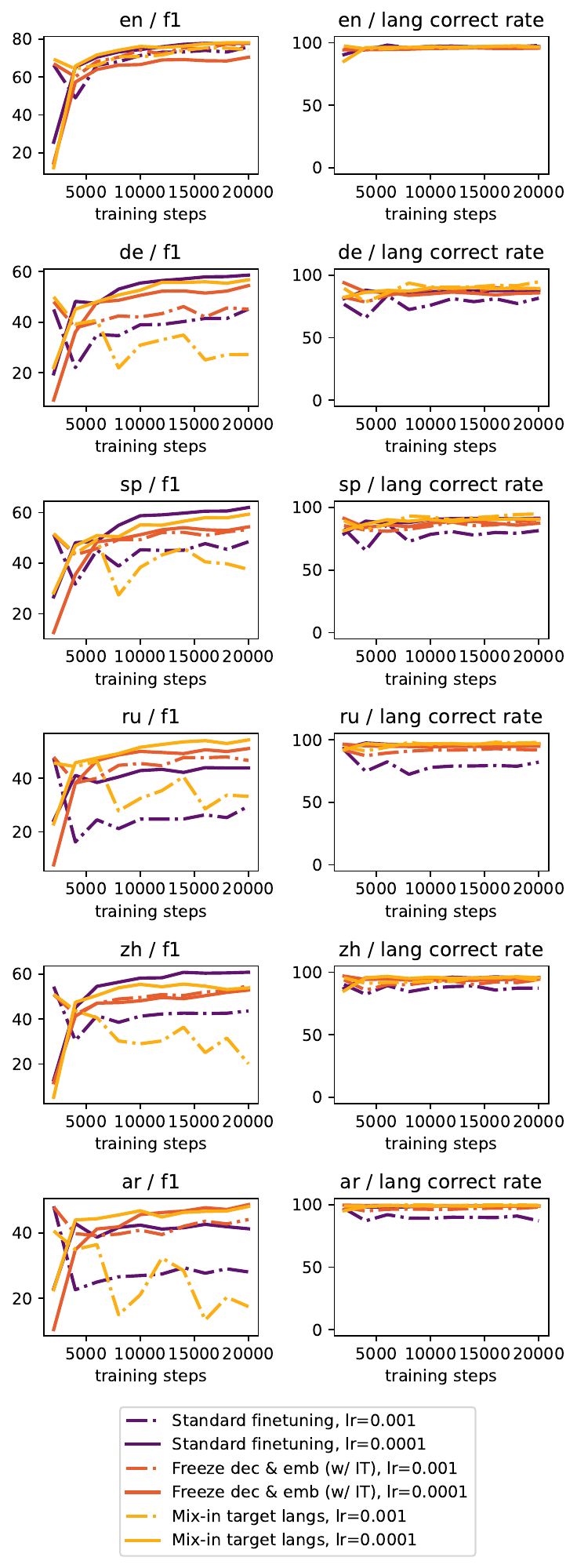} 
    \caption{Per-language results on the effect of learning rate, for mT5 on XQuAD.}
    \label{fig:mt5_xquad}
\end{figure}

\begin{figure}
    \centering
     \includegraphics[width=\linewidth]{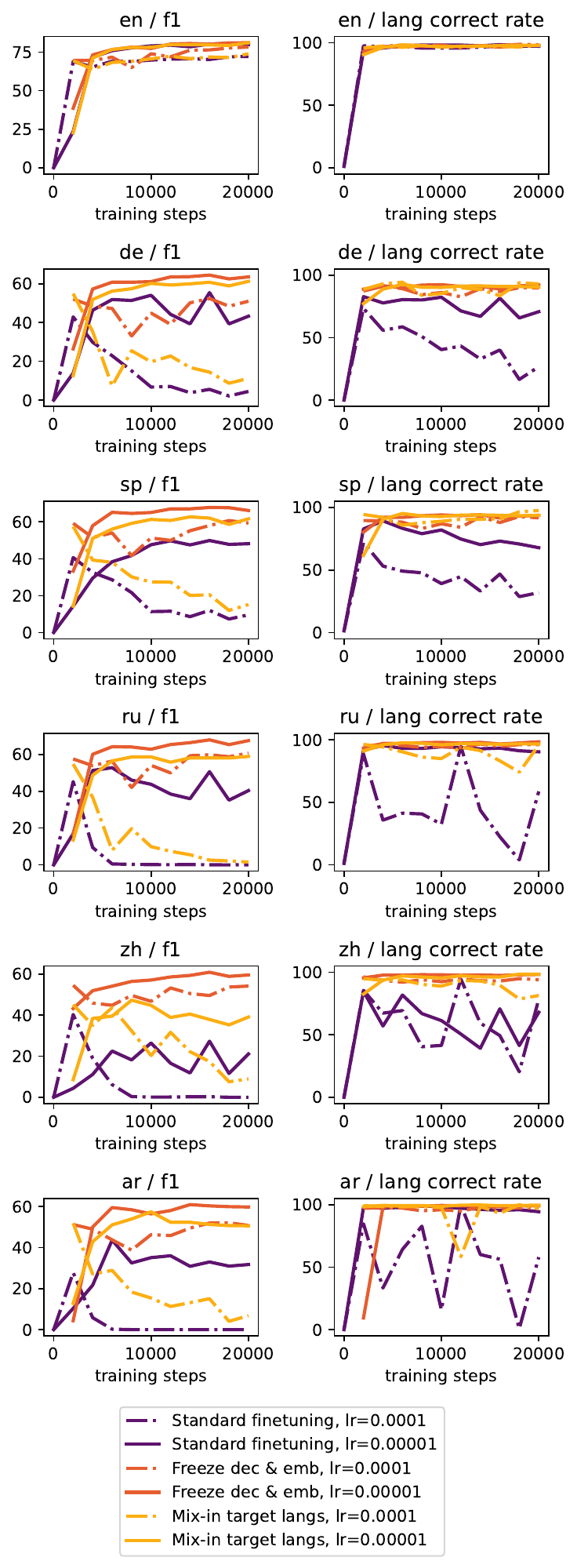}
    \caption{Per-language results on the effect of learning rate, for mBART on XQuAD.}
    \label{fig:mbart_pt_xquad}
\end{figure}

\end{document}